\documentclass[11pt]{article}

\usepackage[preprint]{acl}

\usepackage{times}
\usepackage{latexsym}
\usepackage[T1]{fontenc}
\usepackage[utf8]{inputenc}
\usepackage{microtype}
\usepackage{inconsolata}
\usepackage{graphicx}
\usepackage{booktabs}
\usepackage{xcolor}
\usepackage{amsmath}
\usepackage{amssymb}
\usepackage{multirow}
\usepackage{enumitem}

\usepackage{algorithm,algorithmic}

\title{Capability-Aligned Hierarchical Learning for Tool-Augmented LLMs}

\author{
Haotong Yang \and Ting Long\thanks{Corresponding authors.} \and Yi Chang\footnotemark[1] \\
Jilin University \\
\texttt{haotong25@mails.jlu.edu.cn, longting@jlu.edu.cn, yichang@jlu.edu.cn}
}

\begin{document}
\maketitle
\begin{abstract}
Tool learning enables LLMs to invoke external tools to accomplish tasks. Prior studies have demonstrated the effectiveness of a hierarchical structure: a high-level policy handles global planning and decomposes tasks into manageable sub-tasks, and a low-level policy focuses on invoking tools to solve these sub-tasks. However, these works typically optimize the high-level and low-level policies separately, leading to planner-executor misalignment and limiting LLM performance on tool-use tasks. In this paper, we propose a method called Capability-Aligned Hierarchical Learning (CAHL), which leverages RLVR to jointly optimize both policies, enabling better alignment between the high-level planner and the low-level executor. Experiments on constrained tool-use benchmarks (API-Bank and BFCL) and an open-ended environment (Bamboogle) demonstrate the effectiveness of CAHL. Our code is available at \url{https://github.com/al123123123-123/htl}.

\end{abstract}

\section{Introduction}

Tool learning refers to the ability of large language models (LLMs) to invoke external tools \citep{schick2023toolformer, qin2024tool}, such as APIs \citep{qintoolllm, patil2024gorilla}, web search engines \citep{lazaridou2022internet, jin2025search}, to accomplish tasks that cannot be solved by text generation alone. For example, given the query \textit{What's the weather in Beijing today?}, directly relying on text generation alone would likely produce outdated information, since LLMs cannot know real-time weather from training data. To address this, LLM calls a weather API with the argument (city: \texttt{Beijing}) and incorporates the returned real-time information into its response.

\begin{figure}[t!]
\vskip 0.2in
\centering
\includegraphics[width=1.0\linewidth]{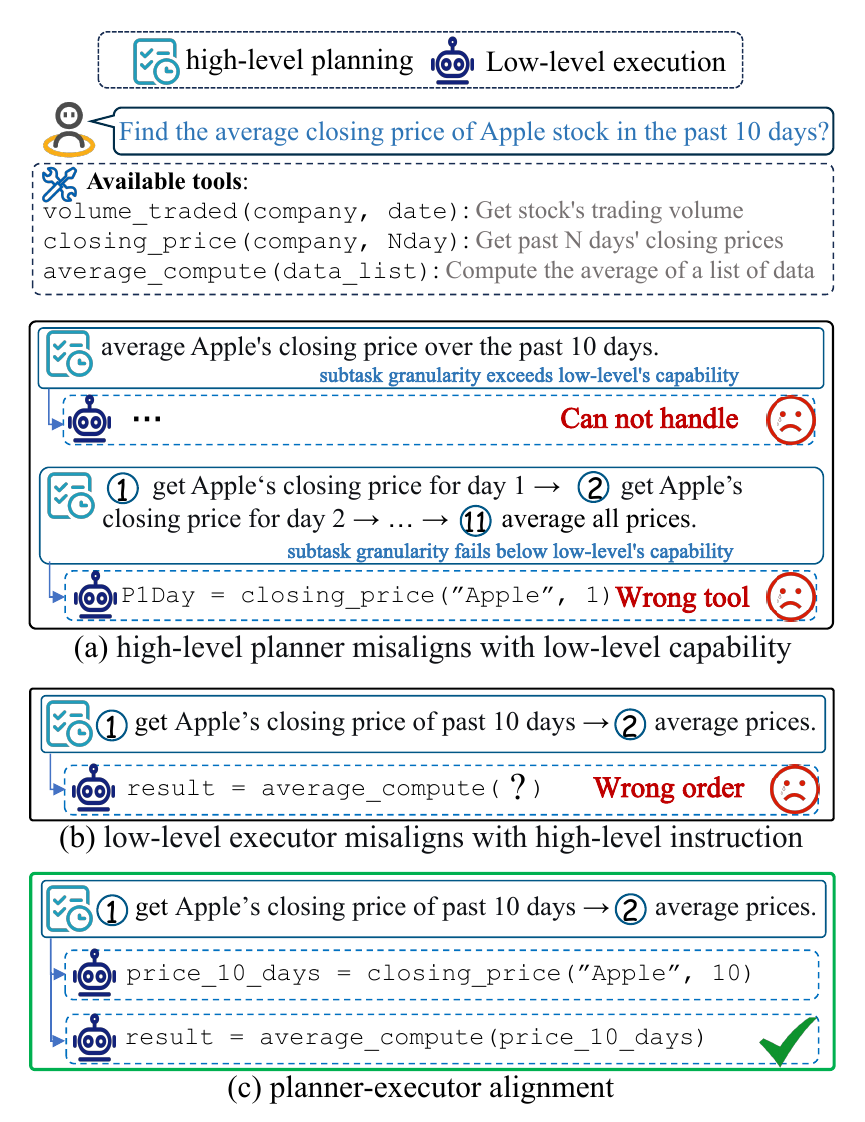}
\caption{
Misalignment between high-level planning and low-level execution. The planner may generate sub-tasks that are logically reasonable but difficult for the low-level executor to realize with available tool schemas, while the executor may fail to align with the planner's underlying intent.
}

\label{fig:intro}
\vspace{-0.5cm}
\end{figure}

However, this single-turn paradigm, where the model performs one tool call, receives its output, and produces a final answer, only suffices for relatively simple queries \citep{yao2022react, li2023api}. In realistic tool-use scenarios, even seemingly simple user requests may require the model to coordinate multiple tools with different functional scopes and parameter requirements. Consider the query in Figure~\ref{fig:intro}: \textit{Find the average closing price of Apple stock in the past ten days}. 
Answering this requires: (1) getting Apple’s closing price for the past 10 days; (2) averaging those prices.
Solving this in a single turn would lead to incorrect or incomplete answers, as a single tool invocation cannot cover the multiple steps required above.

To address that, recent works have adopted hierarchical decision making for tool learning \citep{xu2023rewoo,wang2023plan,prasad2024adapt}. 
A high-level planner first decomposes the user request into a sequence of sub-tasks (e.g., [\texttt{search past 10 days closing price}, \texttt{average the price}]), and a low-level executor then translates each sub-task into concrete tool calls with appropriate parameters. 

However, existing hierarchical methods typically optimize the planner and executor separately: the planner is trained to generate \textit{logically correct} sub-task sequences based on the query, without awareness of whether the executor can actually succeed at those sub-tasks. The executor, in turn, is trained independently to map sub-tasks to tool calls. This independent optimization leads to a \textbf{planner-executor misalignment}: the planner may propose sub-tasks that are theoretically sound but do not match the executor's actual tool-use capability. As shown in Figure~\ref{fig:intro}(a), an overly coarse-grained sub-task cannot be parsed into an executable tool, while an overly fine-grained sub-task may force unnecessary low-level details that are already encapsulated by existing tools. On the other hand,  the executor may fail to align with the planner's underlying intent, as shown in Figure \ref{fig:intro}(b).

To address the planner-executor misalignment, we propose \textbf{Capability-Aligned Hierarchical Learning (CAHL)}. Instead of separately optimizing the planner and the executor, CAHL jointly optimizes them based on Reinforcement Learning with Verifiable Rewards(RLVR). This joint optimization encourages the planner to adapt the granularity of its instructions to the executor's actual capabilities, while enabling the executor to adjust its strategy to match the planner's intent. Extensive experiments on constrained benchmarks (API-Bank, BFCL) and unconstrained open-ended environments (Bamboogle) demonstrate the effectiveness of CAHL. Notably, CAHL achieves consistent improvements over strong single-level and hierarchical baselines across constrained and open-ended tool-use benchmarks.
In summary, our main contributions are as follows:
\begin{itemize}  [leftmargin=1em]

\item We identify planner-executor misalignment in existing hierarchical tool learning methods, where planner-generated sub-tasks do not match the executor's actual tool-use capability.
\item We propose CAHL, a joint optimization framework based on RLVR, which encourages planners and executors to adapt to each other through verifiable task-level feedback.
\item We evaluate CAHL on constrained and open-ended tool-use benchmarks, showing consistent improvements over  baselines.

\end{itemize}

\begin{figure*}[t]
  \vskip 0.2in
\centering
\includegraphics[width=1\textwidth]{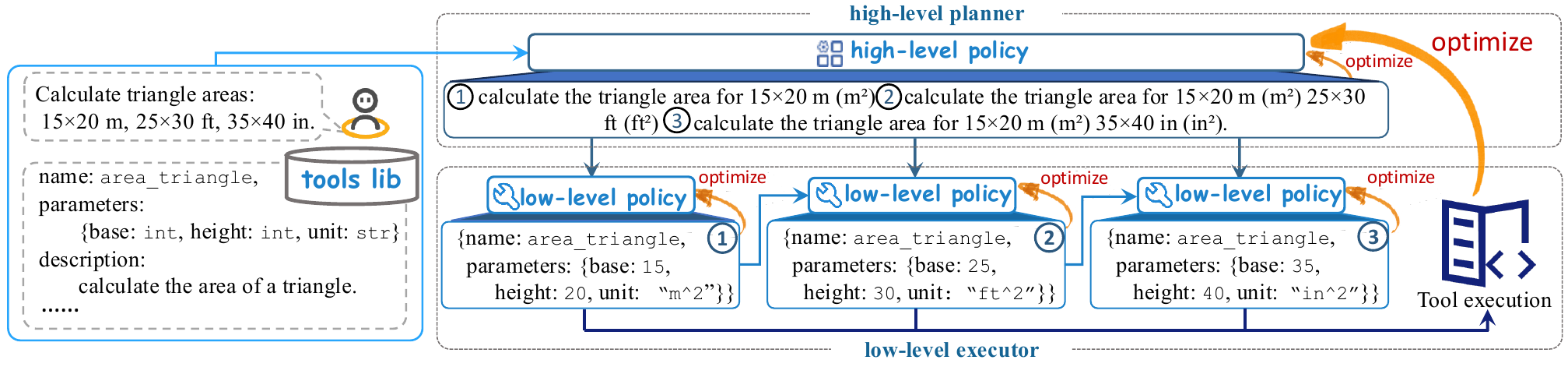} 

\caption{Overview of Capability-Aligned Hierarchical Learning.}
\vspace{-0.5cm}
\label{fig:pipeline}
\end{figure*}

\section{Related Work}
\label{sec:related}

Early explorations of tool learning relied on in-context learning \citep{lazaridou2022internet,paranjape2023art,kim2023language,cummins2025llm,xu2023rewoo,prasad2024adapt}, which supplies few-shot tool invocation exemplars and API specifications within the prompt to guide LLMs in invoking external tools.
With the construction of large-scale instruction datasets such as ToolBench \citep{qintoolllm}, ToolACE \citep{liutoolace}, and xLAM \citep{zhang2025xlam}, supervised fine-tuning (SFT) became the dominant paradigm \citep{patil2024gorilla,acikgoz2025can,chen2024agent}. SFT enhances the adherence to tool invocation schemas and precision in following complex API execution protocols,
but recent work shows that these methods fail to generalize to unseen tools and complex multi-step tasks

\citep{zeng2024agenttuning,chu2025sft,guo2025deepseek}.
More recently, reinforcement learning (RL) has been adopted to optimize tool use via interaction feedback \citep{li2025torl,jin2025search}. ToolRL \citep{qian2025toolrl} introduces fine-grained reward decomposition, separating format correctness from semantic correctness to provide denser signals for optimization. ToolZero \citep{zeng2025tool} further demonstrates that RL, without SFT warm-up, can elicit intrinsic tool-use capabilities directly from base models. In addition, ToolSample \citep{feng2025toolsample} improves sample efficiency by incorporating dynamic task sampling and curriculum learning strategies.

Another line of work studies hierarchical tool use, where a high-level planner decomposes a user request into sub-tasks and a low-level executor maps these sub-tasks into concrete tool calls \citep{xu2023rewoo,wang2023plan,prasad2024adapt}.
This plan-then-execute paradigm improves interpretability and scalability for long-horizon tool-use tasks by separating global task decomposition from local tool execution.
However, most existing hierarchical methods optimize the planner and executor separately.
As a result, the planner may generate sub-tasks that are logically reasonable but difficult for the low-level executor to realize with the available tool schemas. Meanwhile, 
the executor may fail to align with the planner's underlying intent, leading to misordered, incomplete, or redundant tool invocations.

\section{Problem Definition}

In this paper, we study the problem of tool use, where LLMs are required to accomplish user-specified tasks by invoking external tools.
Formally, given a user query $q$ and a predefined set of available tools $\mathcal{T} = \{t_1, t_2, \dots, t_K\}$, the LLM must generate a sequence of actions $\{a_1, a_2, \dots, a_k\}$, where each $a_i$ can be either a text response or a tool invocation.

According to previous works \citep{li2025torl,yu2025steptool,qian2025toolrl,zeng2025itool}, tool-use problem can be naturally formulated as a decision-making problem and defined as a Markov Decision Process (MDP) tuple $\mathcal{M} = (\mathcal{S}, \mathcal{A}, \mathcal{P}, R, \gamma)$.
$\mathcal{S}$ denotes the state space. For a state $s_k$ at step $k$ ($s_k \in \mathcal{S}$), it summarizes the user query $q$ and the interaction history $\{(a_1, o_1), (a_2, o_2), \dots, (a_{k-1}, o_{k-1})\}$, providing the necessary context for the model to select appropriate tools. Here $o_{k}$ denotes the observations returned by the executed tools at step $k$.
$ \mathcal{A}$ denotes the action space. An action $a_k$ corresponds to either invoking a tool $t_k \in \mathcal{T}$ with specified arguments or a text response (e.g., intermediate reasoning or a final answer).

$\mathcal{P}$ denotes the transition function $\mathcal{P}(s_{k+1} \mid s_k, a_k)$, which transfers from state $s_k$ to $s_{k+1}$ when action $a_k$ is executed.
$R$ denotes reward $R(s_k, a_k)$, 
is the feedback on task performance after executing action $a_k$.
$\gamma$ denotes the discount factor, and $\gamma \in (0,1]$ controls the relative importance of immediate versus future rewards.
Under the MDP formulation, given a state $s_k$, 

the objective of tool use task is to learn a policy $\pi(a_k \mid s_k)$ that maps the state $s_k$ to a tool-use action $a_k$, so as to maximize the expected cumulative reward:

\begin{equation}
\max_{\pi} \; \mathbb{E}_{\pi} \left[ \sum_{k=0}^{T} \gamma^{k} \mathcal{R}(s_k, a_k) \right].    
\end{equation}

\section{Methodology}

\label{sec:method}

To address planner-executor misalignment in hierarchical tool learning, we propose \textbf{CAHL} in this paper. As shown in Figure \ref{fig:pipeline}, CAHL is a hierarchical framework with a high-level planner and a low-level executor. Unlike prior work, we jointly optimize both policies using RLVR, encouraging the planner to match its instruction granularity to the executor's capabilities, and the executor to align its strategy with the planner's intent. We next describe the high-level policy, the low-level policy, and their joint optimization.

\subsection{High-level policy}

The high-level policy is designed to synthesize a comprehensive execution plan $\mathcal{Z}$ in a single pass, providing a global context that specifies the overall tool-use structure and intended execution flow. Formally, given the user query $q$, available tool set $\mathcal{T}$, the high
policy $\pi_h$ generates the global plan:
\begin{equation}
    \mathcal{Z} \sim \pi_{h}(\cdot \mid q, \mathcal{T}),
\end{equation}
where $\mathcal{Z}$ denotes the global plan, which is composed of an ordered sequence of $k$ sub-tasks $\mathcal{Z} = [z_1, z_2, \dots, z_k]$. Each sub-task $z_k$ is 
a structured instruction that encapsulates both the tool and the corresponding argument filling: 

\begin{equation}
\small
    z_k = \langle t_k, \{ (p, v_p) \mid p \in \text{Params}(t_k), v_p \leftarrow \text{Extract}(q) \} \rangle,
\end{equation}
where $t_k \in \mathcal{T}$ represents the selected tool for the $k$-th step, and $\text{Params}(t_k)$ denotes the set of required parameters for tool invocation  $t_k$. The value $v_p$ signifies the specific argument value extracted directly from the user query $q$. Notably, the parameters specified by the high-level policy are not fixed: during local execution, they may be adaptively refined based on the actual tool feedback returned at previous steps.

\subsection{Low-level Policy}

The low-level policy is designed to execute individual tool invocation conditioned on the tool specification, intended arguments provided by the high-level policy, and the actual feedback from previous execution steps. It refines the tool arguments when necessary, performs the corresponding tool invocation, and returns the execution result to update the interaction state. Specifically, the low-level policy $\pi_l$ is formulated as:

\begin{equation}
    a_k \sim \pi_{l}(\cdot \mid \mathcal{F}_{1:k}, z_k, \mathcal{T}), 
\end{equation}
where $\mathcal{F}_{1:k}$ denotes the feedbacks of the previous $k - 1$ execution steps.
$a_k$ denotes the final executable output generated by the low-level policy, 
which can be either (1) reasoning text (e.g., intermediate thoughts or a final answer), or (2) a tool invocation formatted as a code snippet or a machine-readable JSON object.

\subsection{Joint Optimization via RLVR}

To align the high-level policy's planning granularity with the low-level policy's execution capabilities, we design a series of verifiable rewards for each policy and jointly optimize both policies with GRPO \citep{shao2024deepseekmath}.

\paragraph{High-level Reward.}

The high-level reward $R_h$ evaluates whether the generated plan leads to accurate parameter inference and correct execution behavior under the given context.
Specifically, the high-level reward is defined as
\begin{equation} \label{eq:high_reward}
R_h =  R_{(h,\text{param})} + R_{(h,\text{exe})},
\end{equation}
where $R_{(h, \text{param})}$ is the parameter-level reward, and $R_{(h, \text{exe})}$ is the execution-level reward.

The parameter-level reward $R_{(h,\text{param})}$ evaluates whether the high-level policy can extract correct argument from the provided context. Suppose the $\mathcal{V}$ denotes the set of all argument values contained within these ground truth tool invocations, the parameter reward is defined by: 
\begin{equation} 
    R_{(h, \text{param})} = \sum_{v_p \in \mathcal{V}} \mathbb{I}(v_p \subseteq \mathcal{Z}),
\end{equation}
where $v_p \subseteq \mathcal{Z}$ denotes that the value string $v_p$ appears as a substring within the generated plan $\mathcal{Z}$. $\mathbb{I}[\cdot]$ is the indicator function, which outputs $1$ when the generated arguments match the expected values (e.g., specific city names or dates) prior to final structured parsing, and $0$ otherwise.

The execution-level reward assesses the correctness of the generated plan based on its induced execution trajectory. Specifically, given a global plan $\mathcal{Z}$, the low-level policy follows this plan and produces an execution trajectory $\tau = \{a_1, a_2, \dots, a_k\}$, where each element $a_j$ corresponds to either a concrete tool invocation or a natural language output at step $j$. The execution-level reward is defined as the average step-wise alignment between the generated execution trajectory and the ground-truth sequence $\mathcal{G} = \{g_1, g_2, \dots, g_n\}$:

\begin{equation}
R_{(h,\text{exe})} = \frac{1}{k} \sum_{j=1}^{k} \text{Score}(a_j, g_j).
\end{equation}

Here $\text{Score}(a_j, g_j)$ is computed based on the alignment between the executed action produced by the low-level policy and the corresponding ground-truth.
If both $a_j$ and $g_j$ correspond to tool invocations, and the executed tool name and all associated argument values in $a_j$ exactly match those specified in $g_j$, $\text{Score}(a_j, g_j)$ is set to $1$;
If both $a_j$ and $g_j$ are natural language responses (e.g., intermediate reasoning traces or final answers), $\text{Score}(a_j, g_j)$ is defined as the semantic similarity between the two outputs. In all other cases, such as action type mismatch or structurally invalid executions, $\text{Score}(a_j, g_j)$ is set to $0$. That is:
\begin{equation}
\small
    \text{Score}(a_j, g_j) = 
\begin{cases} 
\mathbb{I}[a_j = g_j], & \text{if } g_j \in \mathcal{A}_{\text{tool}}, \\
\text{CosSim}(a_j, g_j), & \text{if } g_j \in \mathcal{A}_{\text{text}}, \\
0, & \text{otherwise}.
\end{cases}
\end{equation}

By grounding the high-level reward in the actual execution of the low-level policy, this execution-aware design enables effective feedback from low-level execution outcomes to high-level planning. As a result, the planner is encouraged to generate global plans that are not only structurally valid but also well-aligned with the execution capabilities of the low-level policy and the ultimate tool-use objective.

\paragraph{Low-level Reward.}

Formally, the low-level reward is defined as:
\begin{equation} \label{eq:low_reward}
    R_{l} = R_{(l,\text{form})} + R_{(l,\text{syn})} + R_{(l, \text{sem})},
\end{equation}
where $R_{(l,\mathrm{form})}$ denotes the format-level reward that evaluates structural validity, $R_{(l,\mathrm{syn})}$ denotes the syntax-level reward that assesses schema compliance, and $R_{(l,\mathrm{sem})}$ denotes the semantic-level reward that measures semantic correctness.

The format-level reward is a binary reward, which verifies whether the generated output $a_k$ contains all required special tokens (e.g., XML tags) in the strict order specified by the prompt.

\begin{equation}
R_{(l,\mathrm{form})} = \begin{cases} 1, & \text{all tokens present and ordered},\\ 0, & \text{otherwise}.\end{cases}
\end{equation}

The syntax-level reward evaluates whether the parameters in the generated output $a_k$ strictly adhere to the predefined tool schema, including the correctness of argument structure and compliance with data type constraints. Specifically, for a tool invocation  $t_k$ in $a_k$, we employ a validation function 
to verify if the parameter $p$ of tool invocation  $t_k$ conforms to the structure and data types defined in the tool environment's schema: if no tool call is detected, the reward is 0; If tool calls are present and all extracted arguments pass the schema validation, the reward is 1; otherwise, the reward is -1. That is:

\begin{small}
\begin{equation}
R_{(l, \text{syn})} =
\begin{cases}
0, & \text{if } a_t \text{ contains no tool invocation}, \\[4pt]
1, & \text{if } \forall p \in \text{Params($t_k$)}, p \text{ is valid in schema}, \\[4pt]
-1, & \text{otherwise}.
\end{cases}
\end{equation}
\end{small}

The semantic-level reward, conditioned on valid formatting, measures the semantic alignment between the predicted tool invocation $ \mathcal{T}' = \{t_1, t_2, ..., t_k\}$ and the ground-truth invocation $\mathcal{G} = \{g_1, \dots, g_n\}$. Specifically, we focus on:

(1) Tool name: we measure  whether the intersection over union (IoU) of selected tool names,
\begin{equation}
    r_{n} = \frac{|N_\mathcal{G} \cap N_\mathcal{T}'|}{|N_\mathcal{G} \cup N_\mathcal{T}'|} \in [0, 1],
\end{equation}
where $N_\mathcal{G}$ and $N_\mathcal{T}'$ are the sets of tool names in $\mathcal{G}$ and $\mathcal{T}'$.

(2) Parameter name:  we measure if the correct arguments are identified,
\begin{equation}
r_{p} = \sum_{g_j \in \mathcal{G}} \frac{|\text{Params}(g_j) \cap \text{Params}(t(g_j))|}{|\text{Params}(g_j) \cup \text{Params}(t(g_j))|},   
\end{equation}
where $\text{Params}(\cdot)$ represents the set of parameter names. $t(g_j)$ denotes the tool invocation produced by the model that corresponds to the ground-truth tool invocation $g_j$.

(3) Parameter value: we check whether the argument values are exactly matched,

\begin{equation}
    r_v = \sum_{g_j \in \mathcal{G}}  
          \mathbb{I}\big[\text{Value}_{g_j}[p] = \text{Value}_{t(g_j)}[p]\big],
\end{equation}
where $p$ is the parameter's name.
$\text{Value}_{g_j}[p]$ and $\text{Value}_
{t(g_j)}[p]$ denote the values of parameter $p$ in the ground-truth tool invocation $g_j$ and the model-generated tool invocation $t(g_j)$, respectively.
$\mathbb{I}[\cdot]$ is the indicator function.

\paragraph{Model training.}
We optimize both high-level policy and low-level policy with GRPO \citep{shao2024deepseekmath}. 
Specifically, for each input query $q$, the high-level policy $\pi_h$ samples a group of $n$ candidate global plans, denoted as ${Z} = \{\mathcal{Z}^{(1)}, \dots, \mathcal{Z}^{(n)}\}$. Each plan $\mathcal{Z}^{(i)}$ is parsed into a sequence of subtasks $\mathcal{Z}^{(i)} = [z_1, z_2, \dots, z_k]$. The high-level trainer waits until the low-level executor completes the entire sequence. Upon completion, it evaluates the comprehensive rewards $\{R_h^{(1)}, R_h^{(2)}, ..., R_h^{(n)}\}$ via Eq.~\eqref{eq:high_reward}. The policy $\pi_h$ is then optimized based on the group advantage derived from the reward distribution within $\mathcal{Z}^{(i)}$:
\begin{equation}
\resizebox{0.9\hsize}{!}{
    $\mathcal{L}_\text{h}(\theta_\text{h}) = - \mathbb{E}_{\mathcal{Z}^{(i)} \sim \pi_\text{h}} \Big[ A \left(\mathcal{Z}^{(i)}, R_h^{(i)} \right) \, \log \pi_\text{h} \left(\mathcal{Z}^{(i)} \mid q; \mathcal{T} \right) \Big]$,}
\end{equation}
where $A(\mathcal{Z}^{(i)}, R_h^{(i)})$ denotes the group advantage of $\mathcal{Z}^{(i)}$.

For a task $z_i$ specified by the global plan $\mathcal{Z}$, the low-policy $\pi_l$ sample a group of execution $\mathcal{A}^{i} = \{a_{1}^{(i)}, \dots, a_{m}^{(i)} \}$.
Upon the tool of each generated output executed, we compute the corresponding instant rewards $\{R_l^{(1)}, R_l^{(2)}, ..., R_l^{(m)}\}$ of the  low-level policy via Eq.~\eqref{eq:low_reward}. The low-level policy is then optimized with:

\begin{equation}
\resizebox{1.0\hsize}{!}{
    $ \mathcal{L}_\text{l}(\theta_\text{l}) = - \mathbb{E}_{a_j^{(i)} \sim \pi_\text{l}} \Big[ A \left(a_j^{(i)}, R_l^{(j)} \right) \, \log \pi_\text{l} \left(a_j^{(i)} \mid \mathcal{F}_{1:i}, z_i, \mathcal{T}  \right) \Big] $,
}
\end{equation}
where $A \left(a_j^{(i)}, R_l^{(j)} \right)$ denotes the group advantage of $a_j^{(i)}$.
The training details of CAHL are illustrated in Algorithm \ref{alg:CAHL_grpo}.

\begin{table*}[t!]
\centering
\begingroup
\small
\setlength{\tabcolsep}{3pt}
\renewcommand{\arraystretch}{1.08}
\caption{
Main results across BFCL, API-Bank, and Bamboogle.
Best and second-best performances are highlighted in \textbf{bold} and \underline{underlined}.
In BFCL, \textit{Relevance} evaluates whether the model correctly determines if tool use is required for a given query, while \textit{Irrelevance} measures whether the model can correctly avoid unnecessary tool invocations when no tool is needed.
``--'' indicates that the corresponding Bamboogle result is unavailable because we could not find public checkpoints for Tool-N1 and ToolSample to reproduce the evaluation under the same setting.
}
\label{tab:main_results_combined}
\begin{tabular}{@{}lcccccccc@{}}
\toprule
\textbf{Benchmark} & \textbf{Metric} & \textbf{ Qwen} & \textbf{EASYTool} & \textbf{TUMIX} & \textbf{Tool-N1} & \textbf{ToolRL} & \textbf{ToolSample} & \textbf{CAHL} \\
\midrule
\multirow{6}{*}{\textbf{BFCL}} 
& Overall Acc & 47.68 & 36.39 & 58.52 & 53.91 & 56.96 & \underline{60.25} & \textbf{61.10} \\
& Non-Live AST & 68.98 & 59.50 & 84.36 & 77.57 & 85.21 & \underline{86.42} & \textbf{87.37} \\
& Live Acc & 63.31 & 50.26 & 74.26 & 73.39 & 73.39 & \underline{76.85} & \textbf{78.31} \\
& Multi Turn & 8.88 & 0.75 & 16.71 & 10.00 & 13.25 & \underline{18.50} & \textbf{18.75} \\
& Relevance & 72.22 & \textbf{88.89} & \textbf{88.89} & 77.78 & \underline{83.33} & \underline{83.33} & \underline{83.33} \\
& Irrelevance & 71.93 & 44.00 & \underline{79.16} & 77.85 & 75.87 & 78.44 & \textbf{79.43} \\
\midrule
\multirow{4}{*}{\textbf{API-Bank}} 
& Overall & 58.29 & 54.61 & \underline{73.70} & 60.47 & 61.81 & 64.99 & \textbf{75.54} \\
& Level 1 & 65.4 & 61.15 & \underline{76.19} & 70.7 & 72.2 & 73.2 & \textbf{81.7} \\
& Level 2 & 43.3 & 35.82 & 53.73 & 46.3 & 56.7 & \textbf{65.7} & \underline{64.2} \\
& Level 3 & 44.3 & 44.27 & \textbf{76.34} & 36.6 & 32.8 & 39.7 & \underline{62.6} \\
\midrule
\textbf{Bamboogle} & Accuracy & 69.6 & 68.8 & \underline{73.6}& - & {72.0} & - & \textbf{75.2} \\
\bottomrule
\end{tabular}
\endgroup
\end{table*}

\section{Experiments}
We conduct extensive experiments to evaluate the performance of CAHL, focusing on the following research questions: 
(i) Is CAHL more effective than existing methods (\textbf{RQ1})?
(ii) How does joint optimization affect overall performance (\textbf{RQ2})?
(iii) Can CAHL perform tool use more efficiently and reliably than baseline methods (\textbf{RQ3})?
(iv) Does CAHL effectively mitigate planner-executor misalignment (\textbf{RQ4})?
Additional evaluations, including training convergence and computational cost analysis, are provided in the Appendix \ref{app:reward} and Appendix~\ref{app:cost_analysis}.

\subsection{Training and Evaluation}

\paragraph{Training.} 
Following ToolSample \citep{feng2025toolsample}, we adopt a 4,000-sample composite dataset consisting of ToolACE \citep{liutoolace}, Hammer \citep{lin2024hammer}, and xLAM \citep{zhang2025xlam}.
To support CAHL's joint optimization, we reorganize the original tool-use trajectories into role-specific training instances: atomic execution instances for the executor and trajectory-level planning targets for the planner.
Further details are provided in Appendix~\ref{app:implementation_details}.

\paragraph{Evaluation.} 
To thoroughly assess the effectiveness of CAHL, we evaluate our method on three representative benchmarks: the Berkeley Function Calling Leaderboard (BFCL) \citep{yan2024berkeley} for structural and functional correctness, API-Bank \citep{li2023api} for complex multi-turn interactive tool-use, and Bamboogle \citep{press2023measuring} for unconstrained, multi-hop reasoning. The details of benchmarks are discussed is provided in Appendix \ref{app:exp_details}.

\label{sec:exp}

\subsection{Baselines}
\label{sec:baselines}

We compare CAHL with six representative baselines ranging from foundational models to state-of-the-art tool-use frameworks: 
\textbf{Qwen-2.5-Instruct} \citep{qwen2025qwen25technicalreport}, 
\textbf{Tool-N1} \citep{zhang2025nemotron}, 
\textbf{ToolRL} \citep{qian2025toolrl}, 
\textbf{ToolSample} \citep{feng2025toolsample}, 
\textbf{EASYTool} \citep{yuan2025easytool}, and 
\textbf{TUMIX} \citep{chen2025tumix}. 
Detailed descriptions and architectural configurations for each baseline framework are provided in Appendix \ref{app:baselines}.

\subsection{Main Results (\textbf{RQ1})}
\label{sec:main_results}

Table~\ref{tab:main_results_combined} shows that CAHL achieves the best overall performance on the evaluated benchmarks while remaining competitive across fine-grained metrics, demonstrating its effectiveness in tool-use tasks.

On \textbf{BFCL}, CAHL achieves 61.10\% overall accuracy, with the best performance on Non-Live AST, Live Acc, Multi Turn, and Irrelevance.
This suggests that CAHL improves both schema-level tool invocation and execution reliability.
Although EASYTool and TUMIX obtain higher Relevance scores, Relevance mainly measures the pre-execution decision of whether a tool call is needed, rather than the correctness of the subsequent tool-use trajectory.
In particular, EASYTool achieves high Relevance but much lower Irrelevance, suggesting that it may be more inclined to trigger tool calls even when they are unnecessary.
In contrast, CAHL achieves the best Irrelevance score while also leading on execution-related metrics, indicating that it improves reliable tool execution without encouraging excessive tool invocation.

On \textbf{API-Bank}, CAHL achieves 75.54\% overall accuracy, obtaining the best overall performance, the best Level 1 result and strong Level 3 performance.
Compared with ToolSample, CAHL slightly trails on Level 2 but performs better on both the overall benchmark and Level 3. 
Since ToolSample focuses on improving RL training through dynamic sampling and curriculum learning, while CAHL directly aligns high-level planning with low-level execution through joint optimization,

this suggests that joint optimization is especially beneficial for tool-use tasks involving longer dependency chains and more complex multi-step execution.

CAHL underperform TUMIX on level 3, we assume that is because TUMIX trades substantially more inference turns and tokens for stronger multi-agent deliberation and error correction.

Finally, on \textbf{Bamboogle}, CAHL reaches 75.2\% accuracy, outperforming other baselines.
Since Bamboogle evaluates final-answer correctness in open-ended multi-hop QA with web search tools rather than intermediate tool-call accuracy, this result suggests that CAHL generalizes well to open-ended tasks.

\begin{table}[t!]
\centering
\caption{Ablation results on BFCL and API-Bank.}
\label{tab:ablation_results}
\renewcommand{\arraystretch}{1.08}

\resizebox{\columnwidth}{!}{
\begin{tabular}{@{}lccccc@{}} 
\toprule
\textbf{Metric} & \textbf{Single-level} & \textbf{LowOnly} & \textbf{FreezeLow} & \textbf{FreezeHigh} & \textbf{Ours} \\
\midrule
\multicolumn{6}{c}{\textbf{BFCL}} \\
\midrule
Overall & 56.96 & 59.24 & 50.34& 53.30 & \textbf{61.10} \\
Non-live & 85.21 & 85.19 & 72.81& 76.50 & \textbf{87.37} \\
Live & 73.39 & 76.02 & 65.66& 66.62 & \textbf{78.31} \\
Multi & 13.25 & \textbf{20.52} & 11.37& 15.25 & 18.75 \\
Rel. & \textbf{83.33} & 77.78 & \textbf{83.33} & 72.22 & \textbf{83.33} \\
Irrel. & 75.87 & 67.03 & 71.22 & 77.72 & \textbf{79.43} \\
\midrule
\multicolumn{6}{c}{\textbf{API-Bank}} \\
\midrule
Overall & 61.81 & 73.37 & 67.50 & 73.03 & \textbf{75.54} \\
Level 1 & 72.18 & 79.95 & 71.18 & 80.20 & \textbf{81.70} \\
Level 2 & 56.72 & \textbf{70.15} & 56.72 & 68.66 & 64.18 \\
Level 3 & 32.82 & 54.96 & 61.83 & 53.44 & \textbf{62.60} \\
\bottomrule
\end{tabular}
}
\vspace{-5pt}
\end{table}

\subsection{Ablation Study (\textbf{RQ2})}
\label{sec:ablation}

We evaluate four variants to examine the contribution of joint optimization:
(i) \textbf{Single-level} is a single-tier tool-use agent, i.e., ToolRL;
(ii) \textbf{LowOnly} uses the trained low-level executor alone for inference, without invoking the high-level planner;
(iii) \textbf{FreezeLow} freezes the low-level executor and only optimizes the high-level planner;
and (iv) \textbf{FreezeHigh} freezes the high-level planner and only optimizes the low-level executor.

Results in Table~\ref{tab:ablation_results} provide three main observations: (1) both \textbf{FreezeLow} and \textbf{FreezeHigh} substantially underperform \textbf{CAHL} on both BFCL and API-Bank.
This shows that optimizing only one side of the hierarchy is insufficient: when either the planner or the executor is frozen, the two components cannot fully adapt to each other's capability boundaries.
In contrast, CAHL jointly optimizes both components, enabling the planner to generate more executable sub-tasks and the executor to better follow high-level intent; (2) the frozen variants consistently underperform \textbf{LowOnly}, and they can even fall below \textbf{Single-level} on BFCL.
This indicates that simply introducing a hierarchical structure does not guarantee better tool use.
If only the executor is optimized while the high-level planner is frozen, the executor may be constrained by mismatched high-level guidance, causing it to perform worse than direct single-level inference; if only the planner is optimized while the low-level executor is frozen, the planner cannot benefit from reciprocal executor adaptation and may still generate sub-tasks whose distribution or granularity is difficult for the executor to realize.
Therefore, a jointly optimized high-level planner is necessary; otherwise, hierarchical guidance can limit rather than improve execution performance; (3) \textbf{LowOnly} outperforms \textbf{Single-level} on both benchmarks.
Since \textbf{LowOnly} uses the trained low-level executor alone during inference, this result indicates that our training pipeline also strengthens the executor itself beyond the single-level RL baseline.
Overall, these results show that CAHL's gains do not come merely from reinforcement learning or hierarchical decomposition alone, but from joint optimization that aligns and improves both the planner and the executor.

\subsection{Efficiency and Quality Analysis (\textbf{RQ3})}
\label{sec:efficiency}

\begin{figure}[t]
  \vskip 0.1in
\centering
\includegraphics[width=1.0\linewidth]{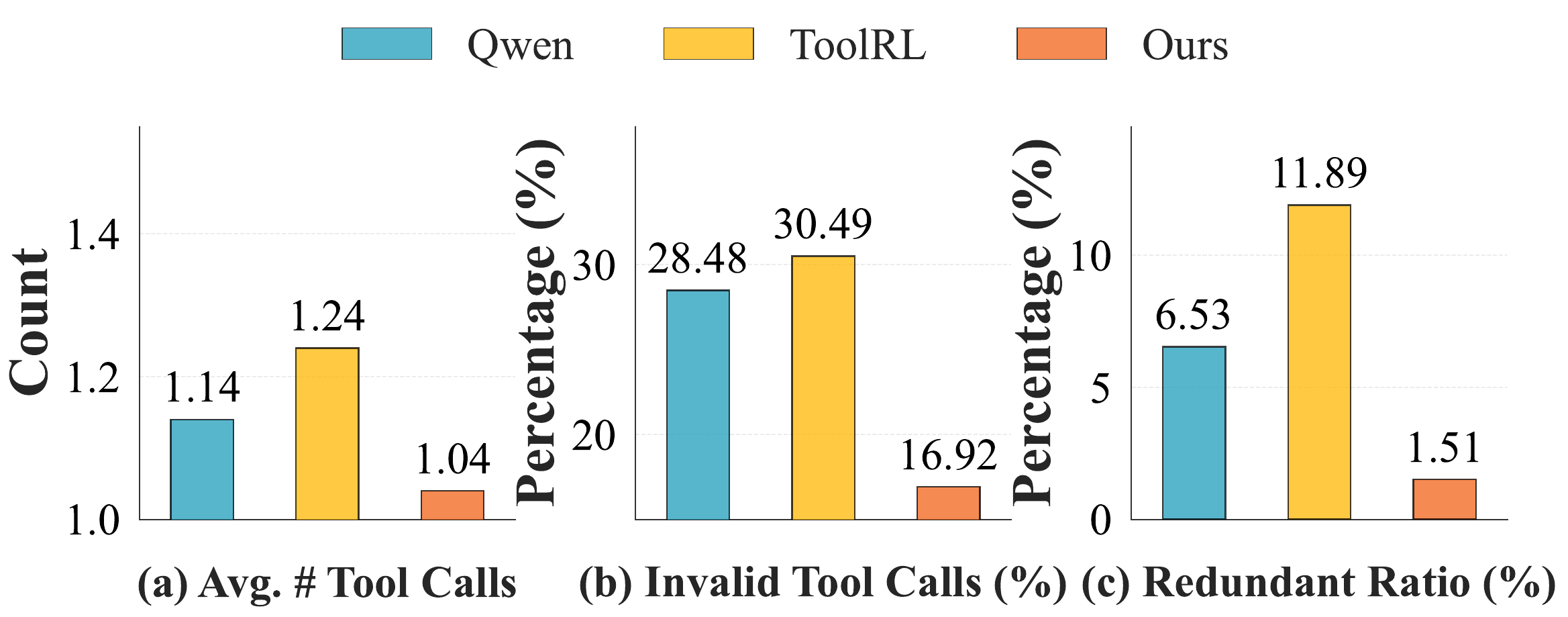}
\caption{Efficiency and quality analysis on API-Bank. CAHL reduces invalid tool invocations and redundant execution loops.} 
\label{fig:efficiency}
\vskip -0.1in
\end{figure}

To evaluate execution efficiency and reliability, we conduct a trajectory-level analysis on 150 randomly sampled single turn  API-Bank tasks without limiting the maximum number of tool calls.
We report three metrics: the average number of tool calls per successful task, the invalid invocation rate, and the redundant call ratio.
Invalid invocations include schema violations, incorrect parameters, and execution errors, while redundant calls refer to repeated or unnecessary tool invocations.
Lower values indicate more concise and reliable tool-use trajectories.

As shown in Figure~\ref{fig:efficiency}, CAHL produces shorter and more reliable execution trajectories:
(1) CAHL requires only 1.04 tool calls per successful task, compared with 1.24 for ToolRL, indicating that it completes tasks with fewer unnecessary actions.
(2) CAHL reduces the invalid invocation rate to 16.92\%, substantially lower than Qwen (28.48\%) and ToolRL (30.49\%), suggesting better schema compliance and parameter grounding.
(3) CAHL lowers the redundant call ratio to 1.51\%, compared with 6.53\% for Qwen and 11.89\% for ToolRL, showing that it is less likely to enter repetitive or ineffective tool-use loops.
Overall, these results indicate that CAHL improves not only final task performance but also the quality and efficiency of intermediate tool execution.

\subsection{Case Study (\textbf{RQ4})}
\label{sec:case_study}

To further analyze how CAHL addresses planner-executor misalignment, we conduct a case study comparing CAHL with a decoupled hierarchical baseline on the same constrained multi-step task.
As shown in Table~\ref{tab:case_study_misalignment}, the decoupled baseline has a planner and an executor, but they are not jointly optimized.
Its unaligned planner issues overly abstract instructions that drop critical constraints (e.g., passwords), while the executor, lacking sufficient high-level guidance, hallucinates missing parameters to satisfy the API schema, leading to execution failure.
In contrast, CAHL jointly optimizes the planner and the executor: the planner anchors required constraints and produces schema-aware guidance, while the executor faithfully grounds this guidance into valid tool calls.
This case highlights the importance of joint optimization for planner-executor alignment, where the planner must adapt its instructions to the executor's executable action space and the executor must reliably realize the planner's intent during tool execution.

\begin{table}[h]
    \centering
    \caption{Case study illustrating the mitigation of planner-executor misalignment. The decoupled baseline generates an overly abstract plan, causing the executor to drop constraints and hallucinate parameters. CAHL enforces strict parameter anchoring via an aligned planner.}
    \label{tab:case_study_misalignment}
    \resizebox{\columnwidth}{!}{
        \renewcommand{\arraystretch}{1.3}
        \scriptsize
        \begin{tabular}{p{\linewidth}}
            \toprule
            \multicolumn{1}{c}{\textbf{Case Study: Mitigating Planner-Executor Misalignment}} \\
            \midrule
            
            \textbf{\textsf{Context \& User Goal:}} \\
            \quad $\bullet$ \textbf{Detailed Goal:} "I have been experiencing migraines. Please update my medical profile to register this new symptom. My username is Jerry and the required password is \textbf{zxcvbn}." \\ 
            \quad \textit{\textbf{Current State:}} Symptom identified. Ready for profile update. \\
            \quad \textit{\textbf{Remaining Task:}} Execute the highly-constrained \texttt{PersonalInfoUpdate} API. \\
            \midrule
            
            \textbf{\textsf{Decoupled Baseline Output:}} \\
            \textbf{\texttt{\textcolor{blue}{[Untuned High-Level Plan]:}}} \\
            \quad \textit{"Subtask: Update the user's medical profile with the new symptom."} \\
            \texttt{\textcolor{orange}{\textbf{<tool\_call>}}} \\
            \{ "name": "PersonalInfoUpdate", \\
            \quad "parameters": \{ "username": "Jerry", "symptom": "migraine", \textbf{\textcolor{brown}{"password": "password123"}} \} \} \\
            \texttt{\textcolor{orange}{\textbf{</tool\_call>}}} \\
            \textbf{\textcolor{brown}{[Critique]: Planner-Executor Misalignment \& Hallucination}} \\
            \textcolor{brown}{$\times$ \textbf{Misalignment:}} The planner issues an overly abstract instruction, dropping the critical password constraint. \\
            \textcolor{brown}{$\times$ \textbf{Hallucination:}} The executor, forced to complete the API schema without guidance, hallucinates a fake password, causing a fatal API error. \\
            \midrule
            
            \textbf{\textsf{Ours (CAHL) Output:}} \\
            \textbf{\texttt{\textcolor{blue}{[Aligned High-Level Plan]:}}} \\
            \quad \textit{"Subtask: Call PersonalInfoUpdate for username 'Jerry' with symptom 'migraine'. Required authentication: password is 'zxcvbn'."} \\
            
            \texttt{\textcolor{violet}{\textbf{<think>}}} Guided by the plan, I will inject the exact credentials... \texttt{\textcolor{violet}{\textbf{</think>}}} \\
            \texttt{\textcolor{orange}{\textbf{<tool\_call>}}} \\
            \{ "name": "PersonalInfoUpdate", \\
            \quad "parameters": \{ "username": "Jerry", "symptom": "migraine", \textbf{\textcolor{teal}{"password": "zxcvbn"}} \} \} \\
            \texttt{\textcolor{orange}{\textbf{</tool\_call>}}} \\
            \textbf{\textcolor{teal}{[Analysis]: Capability-Aligned Execution}} \\
            \textcolor{teal}{$\checkmark$ \textbf{Success:}} The co-optimized planner precisely matches the executor's schema needs, explicitly anchoring the strict constraints and preventing hallucination. \\
            \bottomrule
        \end{tabular}
    } 
\end{table}

\section{Conclusion}

\label{sec:concl}

We propose Capability-Aligned Hierarchical Learning (CAHL) to address planner-executor misalignment in hierarchical tool learning.
CAHL jointly optimizes a high-level planner and a low-level executor with verifiable rewards, encouraging planner-generated sub-tasks to match the executor's actual tool-use capability, while
enabling the executor to adjust its strategy to
match the planner’s intent.
Experiments show that CAHL improves task performance and execution reliability compared with strong baselines.

\section{Limitations}

The hierarchical design of CAHL, while beneficial for complex tasks, introduces additional computational overhead compared to single-level agents. Specifically, the mandatory high-level planning stage before each low-level execution not only increases inference latency but also adds to the overall training cost. This reflects a fundamental trade-off between computational efficiency and execution reliability: the overhead is essentially the price paid for better handling of complex task in  tool usage. Future work could explore more efficient planning strategies as well as asynchronous execution between planning and acting to mitigate this overhead.

\bibliography{custom}

\appendix

\section{Appendix}

\label{sec:Appendix}

\subsection{Detailed Baseline Descriptions}
\label{app:baselines}
To provide a comprehensive benchmark context for resolving planner-executor misalignment, we delineate the architectural and optimization characteristics of the six baselines referenced in Section~\ref{sec:baselines}:
\begin{itemize}[leftmargin=1.5em, itemsep=0.1in]
    \item \textbf{Qwen-2.5-7B-Instruct} \citep{qwen2025qwen25technicalreport} serves as our foundational backbone model. It is heavily instruction-tuned for general prompt-following across multi-turn user intent tracking but operates without targeted parametric enforcement for fine-grained tool schema restrictions.
    \item \textbf{Tool-N1} \citep{zhang2025nemotron} utilizes an early reinforcement learning baseline that optimizes tool operations through sparse binary rewards derived exclusively from final environment execution outcomes, lacking intermediate process-oriented step supervision.
    \item \textbf{ToolRL} \citep{qian2025toolrl} introduces an advanced GRPO-based optimization framework that incorporates dense, structural, and semantic reward parameters to penalize formatting errors during sequential tool invocations. However, it operates purely as a single-level policy under intense long-context input dilution.
    \item \textbf{ToolSample} \citep{feng2025toolsample} enhances ToolRL by introducing dynamic curriculum training paths and specialized sequence difficulty samplers to prioritize complex and borderline execution trajectories during policy adaptation.
    \item \textbf{EASYTool} \citep{yuan2025easytool} structures a prompt-engineered hierarchical workflow that manually orchestrates problem-solving boundaries into disconnected multi-agent planning and grounding steps. It performs inference without joint backpropagation parameter co-optimization.
    \item \textbf{TUMIX} \citep{chen2025tumix} represents a complex multi-agent framework that implements iterative consensus loops to execute test-time computation expansion. While proficient at error correction, its multiple conversation turns incur extreme scaling overhead and latency costs.
\end{itemize}

\subsection{Implementation Details}
\label{app:implementation_details}
We train our CAHL models using the TRL framework\footnote{\url{https://github.com/huggingface/trl}}, a reinforcement learning framework for language models, on 2 NVIDIA A40 GPUs for 15 epochs. To encourage policy exploration, following \citep{wu2025takes}, we generate 2 responses per query in high-level policy training and 4 responses per query in low-level policy training, omit the KL regularization term of GRPO, and set the generation temperature to 1.0. Models are initialized from Qwen-2.5-Instruct-7B and fine-tuned in a parameter-efficient manner using LoRA \citep{hu2022lora} with a rank of 16, with acceleration provided by the Unsloth library\footnote{ \url{https://github.com/unslothai/unsloth}}. The details of our prompts are shown in Appendix Figure \ref{fig:bothprompt}. 

\subsection{Experiment Details}
\label{app:exp_details}
We conduct evaluations on three widely used benchmarks to assess both atomic tool invocation accuracy and long-horizon tool-use performance: BFCL V3 \citep{yan2024berkeley} and API-Bank \citep{li2023api}, we show the statistics of the two benchmarks in table \ref{tab:apibank_stats} and table \ref{tab:bfcl_breakdown}.

API-Bank targets more complex, multi-step tool-use scenarios, where models must interact with multiple APIs over long horizons to complete structured tasks. The benchmark categorizes tasks into three levels based on interaction complexity. We evaluate model performance using overall accuracy and report results separately on Level~1, Level~2, and Level~3, corresponding to single-step API calls, multi-step API interactions with simple dependencies, and long-horizon, multi-API tool-use with complex dependencies, respectively. API-Bank evaluates multi-turn planning and reasoning with an average of 2.91 turns per dialogue, it predominantly features single-call turns (363) over multi-call turns (122).

BFCL V3 focuses on individual tool invocation under strict function schemas, targeting precise argument selection and compliance with the required format. On this benchmark, we report overall accuracy as well as a set of fine-grained accuracy-based metrics that evaluate different invocation scenarios, including non-live AST accuracy (structural and argument-level correctness for non-live calls), live accuracy (execution success for live APIs), multi-Turn accuracy (correctness over multi-turn interactions), relevance detection (identifying when a tool call is required), and irrelevance detection (identifying when no tool call is needed).

Bamboogle is a representative multi-hop question answering benchmark for evaluating free-form tool use in open-ended QA scenarios. Unlike BFCL and API-Bank, it does not evaluate the correctness of intermediate tool invocations or impose strict constraints on tool-call parameters. Instead, models are equipped with a web search tool and evaluated by final answer accuracy. This goal-oriented setting requires the model to decide when to search, interpret retrieved evidence, and synthesize the final answer across multiple reasoning steps.

\begin{table}[htbp]
    \centering
    \begin{minipage}[t]{0.40\textwidth}
        \caption{Statistics of the API-Bank Evaluation Set.}
        \label{tab:apibank_stats}
        \centering
        \begin{small}
        \begin{sc}
            \begin{tabular}{lr}
                \toprule
                Statistic & Value \\
                \midrule
                Domains Covered & 8 \\
                Total APIs & 73 \\
                Total Dialogues & 314 \\
                Total Turns & 914 \\
                Avg. Turns & 2.91 \\
                \midrule
                \textit{Turn Breakdown:} & \\
                \hspace{2mm}Single-Call & 363 \\
                \hspace{2mm}Multi-Call & 122 \\
                \hspace{2mm}Retrieve+Call & 50 \\
                \hspace{2mm}Plan+Retrieve & 50 \\
                \bottomrule
            \end{tabular}
        \end{sc}
        \end{small}
    \end{minipage}
    \hfill
    \begin{minipage}[t]{0.45\textwidth}
        \caption{Breakdown of the BFCL Benchmark.} 
        \label{tab:bfcl_breakdown}
        \centering
        \begin{small}
        \begin{sc}
            \setlength{\tabcolsep}{3pt} 
            \begin{tabular}{llcl}
                \toprule
                Category & Dim. & Count & Purpose \\
                \midrule
                \multirow{2}{*}{Non-Live} 
                    & AST & 1,150 & Syntax \\
                    & Irrel. & 240 & Halluc. \\
                \midrule
                \multirow{3}{*}{Live} 
                    & AST & 1,351 & Exec. \\
                    & Rel. & 18 & Select \\
                    & Irrel. & 882 & Halluc. \\
                \midrule
                Multi-Turn 
                    & Dial. & 800 & Reason \\
                \bottomrule
            \end{tabular}
        \end{sc}
        \end{small}
    \end{minipage}
\end{table}

\subsection{The Training of CAHL.}
\begin{algorithm}[H]
    \caption{The Training of CAHL.}
    \label{alg:CAHL_grpo}
\begin{algorithmic}[1]
    \STATE {\bfseries Input:} User Query $q$, Tool Set $\mathcal{T}$, Policies $\pi_h, \pi_l$
    
    \STATE $Z \leftarrow (\pi_h(\cdot \mid q, \mathcal{T}), n)$
    
    \FOR{$\mathcal{Z}^{(i)}$ {\bfseries in} $Z$}
        \STATE $\mathcal{F}_{1} \leftarrow \{q\}$
        \STATE $[z_1, \dots, z_m] \leftarrow \text{Parse}(\mathcal{Z}^{(i)})$
        
        \FOR{$z_k$ {\bfseries in} $[z_1, \dots, z_m]$}
            \STATE $\mathcal{A}_k \leftarrow (\pi_l(\cdot \mid \mathcal{F}_{1:k}, z_k, \mathcal{T}), m)$
            \STATE Calculate reward $R_l$ according to Eq.~\eqref{eq:low_reward}
            \STATE $\pi_l \leftarrow \text{GRPO Update}(\mathcal{A}_k, R_l)$
            \STATE $a^* \leftarrow \text{SelectBest}(\mathcal{A}_k)$
            \STATE $\mathcal{F}_{1:k} \leftarrow \text{UpdateHistory}(\mathcal{F}_{1:k}, z_k, a^*)$
        \ENDFOR
        \STATE Calculate reward $R_h^{(i)}$ according to Eq.~\eqref{eq:high_reward}
    \ENDFOR
    
    \STATE $\pi_h \leftarrow \text{GRPO Update}(Z, \{R_h^{(i)}\}_{i=1}^{n})$
\end{algorithmic}
\end{algorithm}

\subsection{Reward Analysis}
\label{app:reward}

\begin{figure}[t]
  \vskip 0.2in
\centering
\includegraphics[width=1.0\linewidth]{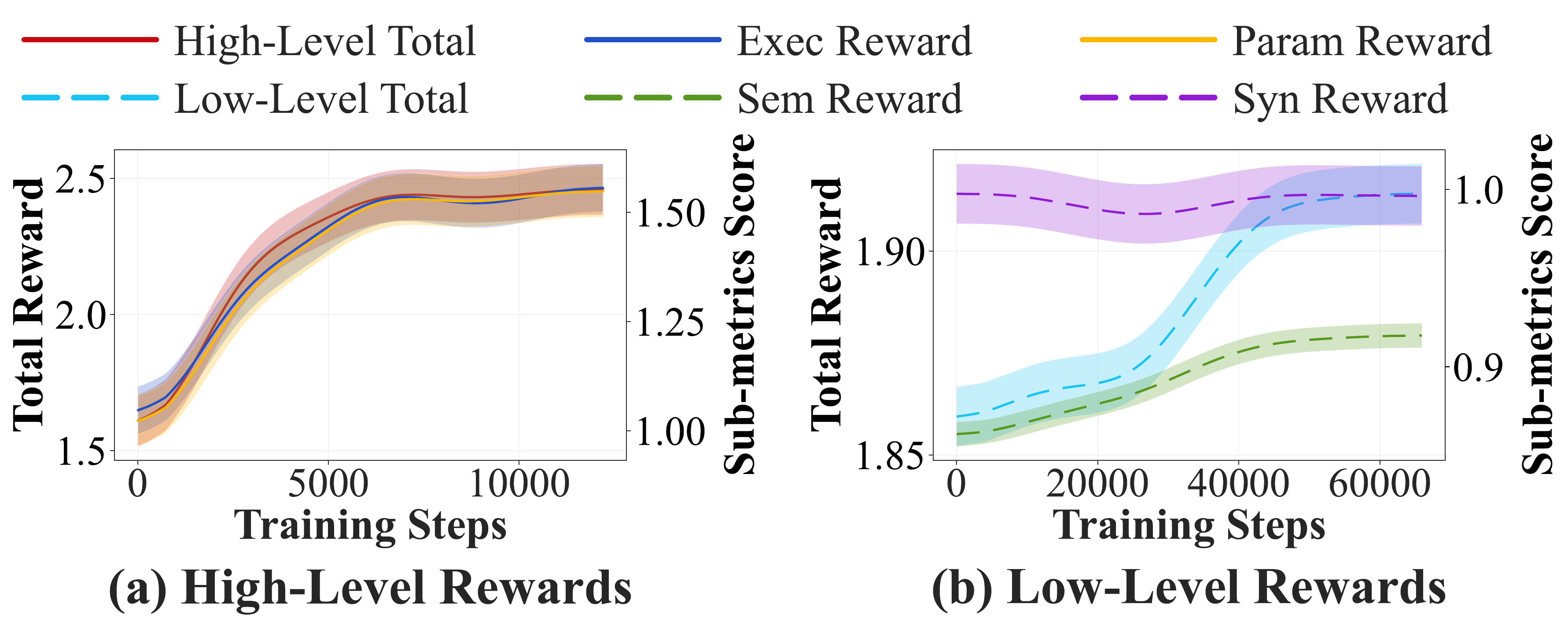}
\caption{The convergence of high-level and low-level rewards.}

\label{fig:curves}
\end{figure}

To analyze how CAHL learns reliable tool use, we visualize the rewards of the high-level planner and the low-level executor during training, as shown in Figure~\ref{fig:curves}.
We observe a coarse-to-fine learning pattern: the high-level policy stabilizes earlier, reaching a relatively high reward within 12,000 steps, while the low-level executor continues to improve over a longer training horizon of about 60,000 steps.

This pattern is consistent with the different roles of the two policies.
The high-level planner mainly learns to produce executable sub-task decompositions with appropriate granularity, while the low-level executor must further learn schema-compliant tool invocation, argument grounding, and feedback-conditioned execution.
Therefore, the early stabilization of the planner provides a more consistent planning distribution, allowing the executor to refine its tool-use behavior under more stable high-level guidance.
This supports the central motivation of CAHL: joint optimization helps the planner and executor gradually adapt to each other's capabilities, rather than optimizing them in isolation.

We also observe a transient decrease in the low-level total reward around 25,000 steps.
This may reflect a temporary adaptation phase, where the executor adjusts to more structured and constraint-aware plans produced by the increasingly stable high-level policy.
After this phase, the low-level reward continues to improve, suggesting that the executor gradually learns to ground high-level intent into executable tool calls.

\begin{table}[h]
    \centering
    \caption{Cost comparison. \textit{Multiple} and \textit{Single} denote multi-turn and single-turn dialogue scenarios, respectively. Both training and testing are performed on an NVIDIA A40 (48GB). Time is measured in seconds (s). \\}
    \label{tab:cost_comparison}
    \resizebox{\linewidth}{!}{
        \begin{tabular}{lcccc}
            \toprule
            
            & \multicolumn{3}{c}{\textbf{Testing}} & \textbf{Training} \\
            
            \cmidrule(lr){2-4} \cmidrule(lr){5-5}

            \textbf{Method} & Multiple (s) & Single (s) & VRAM (GB) & VRAM (GB) \\
            \midrule
            ToolRL & 7.79 & 4.44 & 19.16 & 15 \\
            \textbf{Ours} & 13.21 & 7.21 & 19.50 & 36$^{\dagger}$ \\
            \bottomrule

            \multicolumn{5}{l}{$^{\dagger}$ Includes 21GB (high-level policy) and 15GB (low-level policy).} \\

        \end{tabular}
    }

\end{table}

\subsection{Computational Cost Analysis}
\label{app:cost_analysis}

Although our method achieves strong performance on the benchmark, the hierarchical architecture inevitably introduces additional memory consumption and inference latency.
We compare CAHL with ToolRL in terms of GPU memory usage and inference time, and the results are reported in Table \ref{tab:cost_comparison}. As shown in Table \ref{tab:cost_comparison}, the inference latency for multi-turn tasks increases to 13.21s (compared to 7.79s for ToolRL). This is an expected result of our sequential two-step inference process, which generates a high-level plan before low-level execution. However, thanks to our efficient LoRA implementation, the additional memory overhead during inference is negligible, with VRAM usage remaining comparable to the baseline (19.50GB vs. 19.16GB). In contrast, the training memory consumption approximately doubles (36GB vs. 15GB), reflecting the resource requirements for simultaneously optimizing two policies.
Despite this overhead, the additional cost is a reasonable trade-off for the substantial improvements in execution reliability and reduction of invalid tool calls, which are critical for real-world tool-use applications. Notably, the overhead mainly stems from the hierarchical decision process, and can be further optimized through model compression or context compression strategies. In practice, reliable tool execution is often more critical than marginal inference speedups, making this trade-off acceptable in many real-world scenarios.

\begin{figure*}[t!] 
    \centering
    \includegraphics[width=\textwidth]{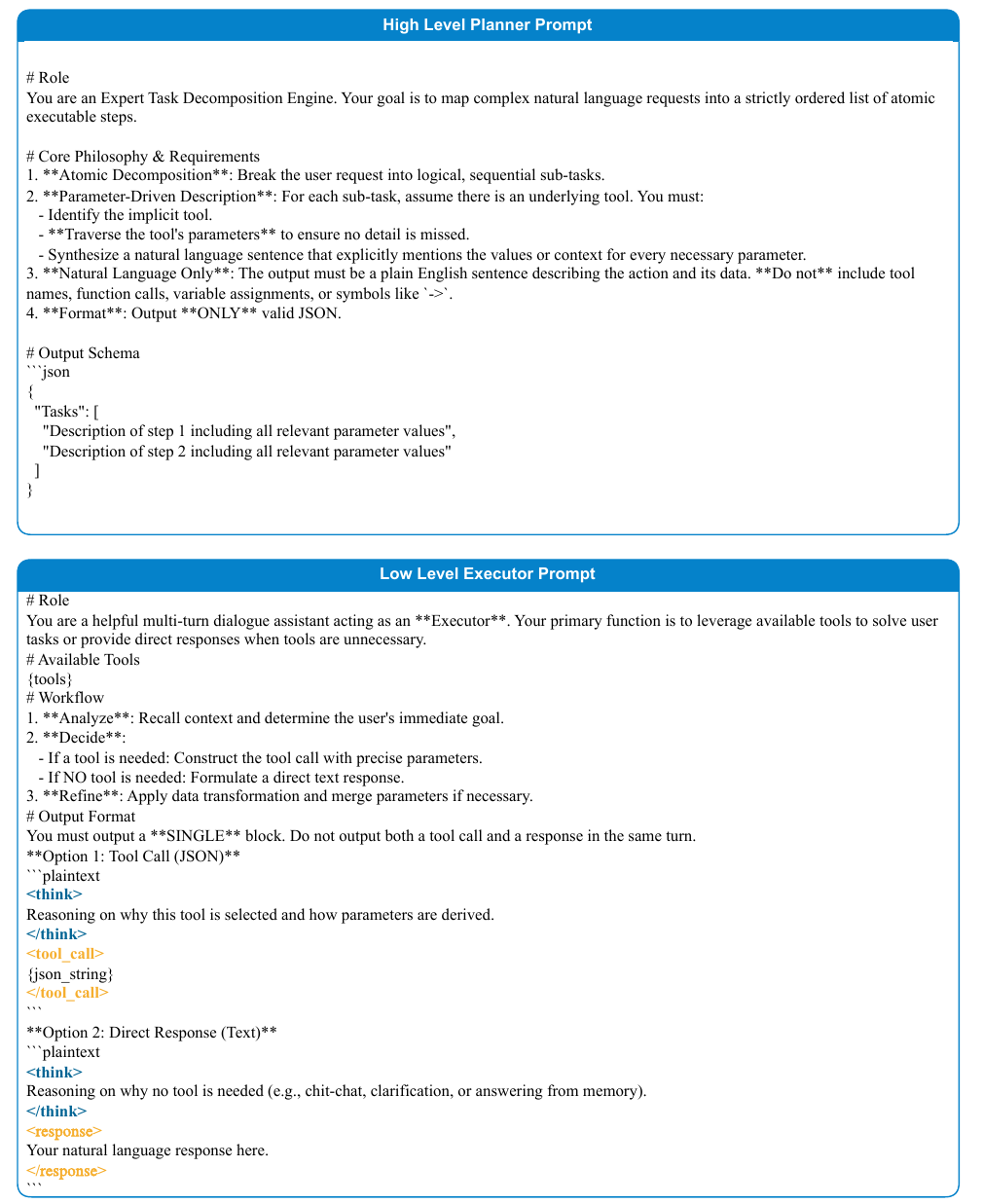} 
    \caption{High-Level Planner Prompt \& Low-Level Executor Prompt.}
    \label{fig:bothprompt}
\end{figure*}

\end{document}